\documentclass{article}

\usepackage[preprint]{template}

\usepackage[utf8]{inputenc} 
\usepackage[T1]{fontenc}    
\usepackage{hyperref}       
\usepackage{url}            
\usepackage{booktabs}       
\usepackage{amsfonts}       
\usepackage{nicefrac}       
\usepackage{microtype}      
\usepackage{xcolor}         

\usepackage{amsmath}

\usepackage[misc]{ifsym}
\usepackage{graphicx}
\usepackage{multirow}
\usepackage{siunitx}

\title{Interpreting Protein Language Model Embeddings via Orthogonal Projection for Protein Fitness Prediction}

\affiliation{hpi}{Hasso Plattner Institute, University of Potsdam}
\affiliation{broad}{Bioinformatics and Machine Learning, Broad Institute of MIT and Harvard}
\affiliation{hub}{School of Business and Economics, Humboldt University of Berlin}
\affiliation{kit}{Scientific Computing Center, Karlsruhe Institute of Technology}

\author{Paulo Yanez Sarmiento}{hpi}
\author{Pia Francesca Rissom}{hpi,broad}
\author{Manuel Pfeuffer}{hub}
\author{Marco Simnacher}{hub}
\author{Jordan F. Safer}{broad}
\author{Sumaiya Iqbal}{broad}
\author{Henrike O. Heyne}{hpi}
\author{Nadja Klein}{kit}
\author{Bernhard Y. Renard}{hpi}

\email{hpi.de}{paulo.yanez}
\email{hpi.de}{francesca.rissom}
\email{hpi.de}{henrike.heyne}
\email{hpi.de}{bernhard.renard}
\email{hu-berlin.de}{manuel.pfeuffer}
\email{hu-berlin.de}{marco.simnacher}
\email{broadinstitute.org}{sumaiya}
\email{broadinstitute.org}{jsafer}
\email{kit.edu}{nadja.klein}

\begin{document}

\maketitle

\begin{abstract}
  Recently, there has been a growing adoption of protein language models (PLMs) in biomedical science. Their embeddings provide a rich numerical representation of protein sequences which achieve state-of-the-art performance on several downstream tasks including protein fitness prediction. However, PLM embeddings are not directly interpretable and, thereby, it remains unclear what features they encode. To gain insight into which biochemical properties of the protein are driving the prediction, we leverage an orthogonal projection technique that removes linear effects of known tabular features from embeddings and extend it to high-order and interaction effects. In this way, we remove the effects of interpretable biochemical features from PLM embeddings. In an ablation study, we show that this leads to a decrease in performance for a downstream classifier trained only on the embeddings to predict protein fitness. In an additional evaluation, we find that these biochemical features explain a substantial part of the variance in the predictions of this classifier. Hence, we can show that PLM embeddings encode patterns correlated with biochemical properties and quantify their contribution to predicting protein fitness. This computationally efficient approach is not limited to the features or embeddings considered here and is readily transferable to problem settings beyond protein fitness prediction.
\\

\textbf{Keywords:} explainable artificial intelligence (XAI) $\cdot$ interpretability $\cdot$ protein language model (PLM) $\cdot$ embedding $\cdot$ orthogonal projection $\cdot$ protein fitness
\end{abstract}

\section{Introduction}\label{sec_introsuction}
Proteins are sequences of amino acids that form large, complex molecules present in virtually all living organisms. They are essential for the structure, function, regulation and communication of cells. Changes in their sequence, such as single substitutions of amino acids (also called variants or mutations), can alter the functional performance (also referred to as fitness) of the protein. As this can be harmful to the organism, it is important to understand the impact of these variants on protein function. For example, in the context of human genetic diseases, it is a central task to correctly predict the functional impact of mutations as the majority of genetic variants in clinical variant databases is still classified as `unknown significance' \cite{rehm2023landscape}. Measuring the impact experimentally can be expensive and time-consuming. Therefore, computational methods have been developed \cite{livesey2025guidelines}. The task of protein fitness prediction refers to estimating how  well a protein performs a desired function relative to a known reference sequence. It is closely related to variant effect prediction which approaches the task more from clinical or mutational perspective \cite{notin2024proteingym}. Among other factors, structural information about the protein has been shown to improve performance for variant effect or protein fitness prediction \cite{gerasimavicius2025leveraging,livesey2025guidelines}.

In contrast to methods relying on engineered features, there has been a growing adoption of protein language models (PLMs) as they provide rich representations of protein sequences. Similar to large language models in natural language processing, PLMs are trained in a task-agnostic, self-supervised manner on large datasets of protein sequences \cite{elnaggar2021prottrans, lin2023evolutionary,rives2021biological, valentini2023promises}. Models trained on PLM embeddings---numerical vectors derived only from the 1D protein sequence---have been shown to achieve state-of-the-art performance on a range of downstream tasks compared to traditional feature-based approaches or methods aligning multiple sequences \cite{derbel2023accurate, marquet2022embeddings,rives2021biological, schmirler2024fine}. However, high-dimensional embeddings are not directly interpretable. Hence, by fitting a downstream classifier on the embeddings, it is unclear which human-understandable features  drive the prediction. Yet, interpretability  is crucial for deriving biological insights and is commonly required for medical applications \cite{vellido2020importance}.

To address this challenge, we leverage an orthogonal projection technique named post-hoc orthogonalization (PHO) that is used in training of semi-structure neural networks (SSNs) \cite{rugamer2023new}. PHO was already applied to biomedical data by \cite{weber2025preventing}, who removed confounding features from radiograph embeddings. Since PHO with respect to tabular features only removes linear effects, we extend the approach to higher-order and interaction effects. Thereby, we provide a computationally efficient approach to help interpret PLM embeddings.

\section{Related Work}\label{sec_related_work}

\paragraph{Interpretation of PLMs} When introducing the PLMs ESM-1 \cite{rives2021biological} and ProtT5 \cite{elnaggar2021prottrans}, the authors showed that the embeddings capture features about structure and biochemical properties by visual analysis after dimensionality reduction.
\cite{vig2021bertology} and \cite{rao_transformer_2020} extended these findings by studying the PLM's Transformer attention maps and leveraged them for contact prediction of amino acids in the protein's 3D structure. \cite{rissom2024decoding} introduced a framework to analyze embedding spaces by considering local neighborhoods and distributions of biological features in the spaces. They demonstrate that by doing this they gain similar insights in what features are encoded in the embeddings as a supervised trained downstream classifier. 
\cite{simon2024interplm} trained sparse autoencoders on PLM embeddings to extract interpretable features from them. Thereby, they could match latent features with known biological concepts relevant for protein binding, structure, and function.

\paragraph{PLMs for Protein Fitness Prediction} Many approaches based on PLMs or their embeddings have been proposed for protein fitness (or variant effect) prediction. In a zero-shot setting, the predicted probability of observing the substituted amino acids at this particular position compared to the non-mutated reference is leveraged for prediction \cite{brandes2023genome,meier2021language, notin2024proteingym}. Alternatively, the embeddings can be used for supervised training of a downstream classifier for regression or classification \cite{derbel2023accurate, marquet2022embeddings,notin2024proteingym}. In the regression case, the model is fitted to a continuous physical quantity which was measured experimentally. For classification, in general, there is a binary distinction between a protein being fit (performing its function) or not. \cite{hsu2022learning} augmented the embeddings with additional site-specific amino acid features for prediction. Furthermore, \cite{schmirler2024fine} showed that task-specific fine-tuning of PLMs mostly increased performance of methods using embeddings as input.

\paragraph{Orthogonal Projection of Features} SSNs have been introduced in the context of distributional regression to overcome the identifiability issue when combining tabular features and latent representation of a deep neural network into a single model \cite{rugamer2023semi}. PHO has been proposed by \cite{rugamer2023new} to train SSNs more effectively and resolve limitations at inference. As mentioned in the introduction, \cite{weber2025preventing} applied PHO to remove biases from radiograph embeddings. In a similar manner, but not in the context of SSNs, \cite{lu2021metadata} remove metadata from neural network embeddings during training via an orthogonal projection to correct for confounding effects. \cite{kohler2024achieving} used PHO to decompose a black-box model into a neural additive model, thereby ensuring a certain orthogonality between higher- and lower-order effects.

\section{Method}\label{section_adaption_attnlrp}

\begin{figure*}
  \centering
  \includegraphics[width=\linewidth]{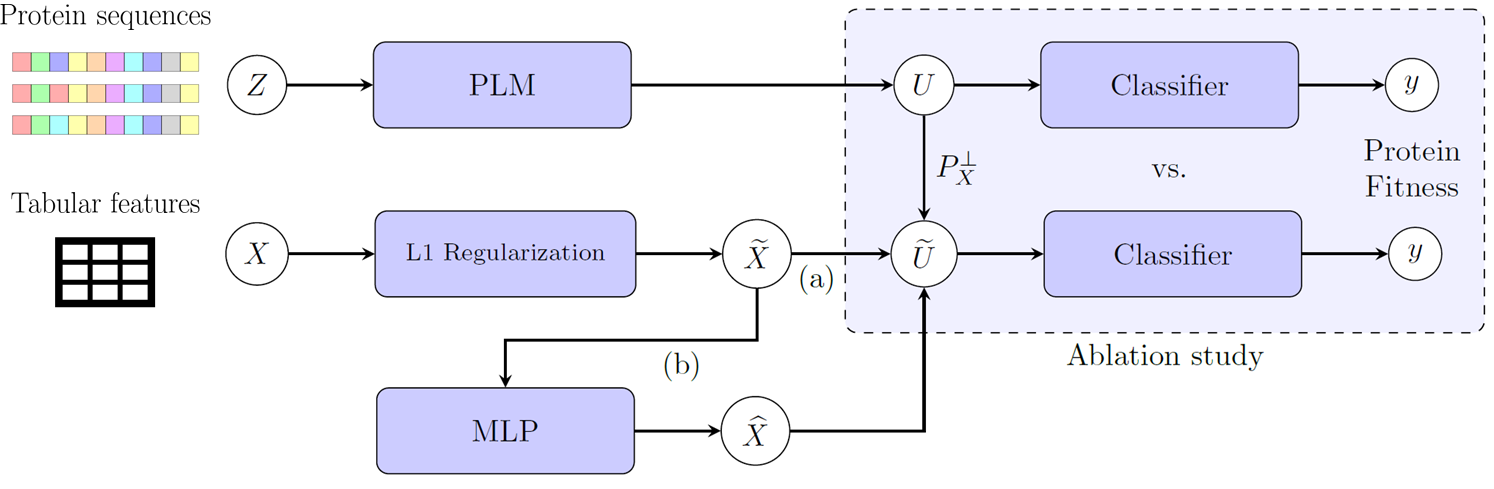}
  \caption{Overview of methodology: Two data modalities are considered: protein sequences $Z$ and corresponding tabular biochemical features $X$. Every data point refers to a variant (or mutation), i.e., where a single amino acid has been substituted compared to the reference protein (or wild-type). The sequences are embedded via a protein language model (PLM), i.e., mapped onto a numerical vector $u\in\mathbb{R}^d$. For the tabular features, a feature selection via $L1$ regularization is applied. These selected features $\widetilde{X}$ are either (a) directly used to project the embeddings $U$ onto their orthogonal complement or (b) a multi-layer perceptron (MLP) is trained on these features and its hidden layer representation $\widehat{X}$ is used for the orthogonal projection $P^\perp_X$. A logistic regression as downstream classifier for protein fitness is trained separately on the embeddings $U$ and the orthogonally projected embeddings $\widetilde{U}$. In an ablation study their predictive performance is compared.
  }
  \label{fig_1}
\end{figure*}

We formulate protein fitness prediction as a binary classification, where $y\in\{0,1\}$, and 1 corresponds to a functioning protein and 0 otherwise. A protein sequence $z$ with maximal length $l$ is mapped to a $d$-dimensional vector by the PLM: $f_{PLM}: \mathcal{A}^l\rightarrow\mathbb{R}^d, z\mapsto u$, where $\mathcal{A}$ is the alphabet of the one-letter codes for the 20 canonical amino acids. The embeddings of $n$ samples of sequences is denoted by $U\in\mathbb{R}^{n\times d}$, while by $X\in\mathbb{R}^{n\times p}$ we denote the corresponding values of  $p$ tabular features.
For our approach to interpret the PLM embeddings, we analyze whether they contain information relevant for protein fitness prediction correlated with the tabular features using an orthogonal projection. 
This analysis consists of multiple steps (see Figure \ref{fig_1}). First, we perform a feature selection to get a subset of tabular features relevant for protein fitness prediction. The selected features are referred to as $\widetilde{X}\in\mathbb{R}^{n\times\widetilde{p}}, \,\widetilde{p}\le p$.
Next, we compare the performance of linear downstream classifiers trained on different data types: the selected tabular features $\widetilde{X}$, a representation of those features denoted by $\widehat{X}\in\mathbb{R}^{n\times\widehat{p}}$, the PLM embeddings $U$, and their respective concatenations $(\widetilde{X},U)$ and $(\widehat{X},U)$ to predict $y$. The representation $\widehat{X}$ is obtained from a hidden layer of a multilayer perceptron (MLP) trained to predict protein fitness on $\widetilde{X}$. This comparison is intended to investigate whether tabular features and embeddings complement each other to predict $y$. 
To test whether embeddings (or tabular features) contain significant information 
for protein fitness beyond the tabular features (or embeddings), we perform conditional independence tests (CITs) \cite{lundborg2024projected, simnacher2026deep}. Specifically, we test $H_0:U\perp y\mid X$ vs. $H_1:U\not\perp y\mid X$ and second $H_0:X\perp y\mid U$ vs. $H_1:X\not\perp y\mid U$, using the algorithm-agnostic, prediction-based CITs from \cite{lundborg2024projected}. 
This means that we test whether the embeddings and protein fitness are independent after conditioning on the tabular features and vice versa.

The key step of our analysis consists of an ablation study in which we quantify the change in performance of the downstream classifier when the effects of the tabular features are removed from the embeddings via orthogonal projection. 
The idea is that if patterns correlated with biochemical features relevant for protein fitness prediction are encoded in the embeddings, their removal should lead to a decrease in performance. We consider both: an orthogonal projection w.r.t. the tabular features and alternatively w.r.t. to their MLP representation to also remove higher-order and interaction effects. 
For readability, in the following we drop $\mathord{\lower.9ex\hbox{$\widetilde{\phantom{x}}$}}$ and $\mathord{\lower.9ex\hbox{$\widehat{\phantom{x}}$}}$ writing only $X$, $x$, and $p$. 
We incorporate a bias term, i.e.,  we consider $[\boldsymbol{1}\quad X]\in\mathbb{R}^{n\times (p+1)}$. However, again for readability we only write $X$ or $x$. The projection onto the orthogonal complement can be rewritten as
\begin{align}
    P_X^\perp U &= \left(I-XX^+\right)U \label{eq_projection_matrix_multiplication} \\
&=U-XD \,, \label{eq_matrix_D}
\end{align}
with Moore-Penrose inverse $X^+=\left(X^\top X\right)^{-1}X^\top$ and matrix $D:=X^+U\in\mathbb{R}^{(p+1)\times d}$. We estimate $D$ given $X^{train}, U^{train}$ and remove the linear effects of $x$ at inference or for our ablation study via $\widetilde{u}= u-x^\top D$. Hence, while the matrix multiplication on the left hand side of \eqref{eq_projection_matrix_multiplication} requires $n$ data points, the matrix $D$ allows to remove the effects sample-wise. We denote the orthogonally projected embeddings by $\widetilde{U}$. For the ablation study, we fit a separate linear downstream classifier on $\widetilde{U}$ and compare its performance to the classifier trained on $U$. 
The approach is computationally efficient because the matrix $D$ must be calculated only once, which involves inverting a $p\times p$ matrix. 

Note that in general the orthogonal projection maps onto a lower-dimensional subspace, i.e., $\dim(\textrm{span} (P_X^\perp U))\le \dim(\textrm{span} (U))$. Hence, to control for the effect of random removal of information in the ablation study, we also calculate an orthogonal projection for a random subspace. For this purpose, the values of the selected tabular features are shuffled in each fold for both the training and test set. This preserves the marginal distributions of the features $\widehat{X}$ while randomizing any correlations among them and with the output variable $y$. Analogously, we also shuffle the values of every dimension in the representation of the tabular features $\widetilde{X}$. With these randomized tabular features or randomized representations, respectively, we also calculate the matrix $D$ from equation \eqref{eq_matrix_D}. In the ablation study, we compare the impact of removing random features against the removal of effects of the actual biochemical features.

In an additional analysis following the approach of \cite{weber2025preventing}, we evaluate how much of the variance in the embeddings-based predictions $\hat{y}_{emb}=\sigma\left(u^\top\gamma\right)$ can be explained with the selected tabular features or their representation, respectively. Therefore, we fit a separate linear model with the predicted logits $\hat{\eta}_{emb}=\sigma^{-1}\left(\hat{y}_{emb}\right)$ as dependent variable
\begin{align}
    \hat{\eta}_{emb} = x^\top\theta \,,
\end{align}
and report its adjusted $R^2$. This model is referred to as \textit{evaluation model}. We perform this evaluation on the hold-out test set $X^{test}, U^{test}$ that was not used for feature selection or fitting $\gamma$.

The code for our implementation and the experiments are available under \url{https://gitlab.com/dacs-hpi/interpret_plms}

\section{Experiments}

\subsection{Data}\label{subsection_data}

We perform our experiments on a subset of ProteinGym \cite{notin2024proteingym}, a collection of large-scale benchmark datasets for protein fitness prediction. Specifically, we focus on deep mutational scanning assays in which protein fitness is experimentally measured in high-throughput for variants of a certain reference protein, also called wild-type. We train the models across assays. Thus, we aim to find features that are relevant for fitness prediction for different proteins. Since the experimental measurements are often referring to different physical quantities across assays and are therefore not directly comparable, we use the binarized score of ProteinGym. Additionally, to maintain consistency among the labels, we split the data by assay function type, i.e., organismal fitness, activity, expression, and binding, and perform the analysis separately on these subsets. We focus on proteins from the human proteome as for those we collect additional tabular features. We also do not consider sequences with multiple mutations, i.e., where multiple amino acids have been substituted. This is because in our approach the tabular features refer to a single mutation. Several assays do not derive from the reference sequence in the UniProt database \cite{uniprot2025uniprot} , or seem to consider only a subsequence. Since our tabular features use this reference, these assays are also excluded. After filtering, we end up with 113k, 69k, 51k, and 18k samples of mutated sequences for the labels \textit{Organismal Fitness}, \textit{Activity}, \textit{Expression}, and \textit{Binding}, respectively with positive-negative class ratios of 1.28, 1.54, 1.63, and 1.23. The samples in the subsets derive from 24, 15, 13, or 6 different reference wild-types, respectively. 
On these subsets, we perform a stratified 5-fold cross validation for our analysis. Note that ProteinGym also contains assays with function type \textit{Stability}. However, after our filtering, only 796 samples of mutated sequences remain. Due to this small sample size, they are not considered in our analysis.

The tabular features are extracted from the Genomics 2 Proteins portal (G2P) \cite{kwon2024genomics}. They consist of structural properties based on the predicted structure for the wild-type sequence from AlphaFoldDB \cite{varadi2024alphafold}. This includes the phi and psi angle of the mutated amino acid (angles between atoms in the amino acid structure), its relative solvent accessibility (RSA), a secondary structure classification, and AlphaFold 2's plddt (predicted local distance difference test) score measuring model confidence \cite{jumper2021highly}. Further, we consider binary features indicating the type of substitution per amino acid, e.g. alanine to aspartic acid, or per chemical property, e.g. aliphatic to nonpolar. Note that the substitution features as well as the secondary structure classification features are mutually exclusive.

\begin{figure}
  \centering
  \includegraphics[width=\linewidth]{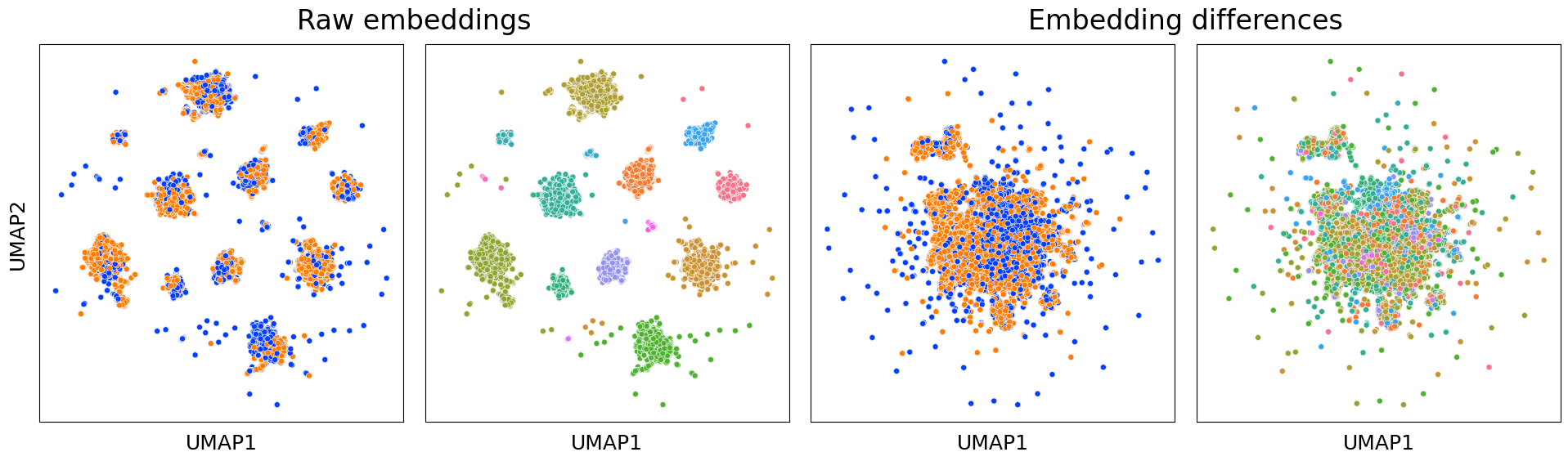}
    \caption{
      Raw embeddings (left) of ESM-2 150M and and their differences to the corresponding reference wild-type embedding (right) after UMAP dimensionality reduction for the subset \textit{Activity}. In the first row, data points are colored based on protein fitness label (positive: orange, negative: blue). The second row displays the same data but the data points are colored based on their corresponding reference wild-type. The raw embeddings cluster by their reference wild-type, the embedding differences are centered.
    }
    \label{fig_umap}
\end{figure}

Given this combination of features describing the biochemical property of the substitution and the structure of the wild-type, we address protein fitness prediction by considering the change from wild-type to variant and not just the variant itself. 
Therefore, we calculate the differences of the embeddings and use them as input for the downstream model, i.e., if $u_0$ denotes the corresponding wild-type embedding, we use $u-u_0$. However, for readability, we only write $u$ or $U$. Figure \ref{fig_umap} illustrates the embeddings and the corresponding differences after dimensionality reduction using UMAP \cite{mcinnes2018umap} exemplary for the subset \textit{Activity}. The dimensionality reduction was only applied for visualization and is not part of the further analysis. We see that while the raw embeddings cluster by their reference wild-type, the embedding differences are centered. Hence, since we consider variants deriving from different wild-types, i.e., across assays (see Section \ref{subsection_data}), the centered data is also more suitable to mitigate bias from a specific wild-type or assay. \cite{gereben2025plm} also showed that considering differences could improve predictive performance.

\subsection{Models}

For our multi-step analysis, we consider different model architectures.

\paragraph{Feature Selection and Representation} We fit a logistic regression model with $L1$ regularization for protein fitness prediction to select relevant tabular features. To balance performance and sparsity, we choose the sparsest model within one standard error of the best-performing model (1-SE rule) by cross-validation (nested in the outer CV). 

For a feature representation that is capable of incorporating higher-order and interaction effects of the tabular features, we train a MLP for protein fitness prediction with two hidden layers of dimension $\hat{p}=\,$256 as a baseline architecture on the selected tabular features $\widetilde{X}$. We also conduct a small hyperparameter search, i.e., scale up width and depth to ensure the model size appropriate for the task and train with different learning rates and dropout.

\paragraph{PLMs} We embed the protein sequences via different Transformer-based PLMs from the widely used ESM family: ESM-1v \cite{meier2021language} and ESM-2 \cite{rives2021biological}. For ESM-2 we consider three different models sizes: 35M, 150M, and 650M parameters corresponding to different embedding dimensions $d=\,$480, 640, 1280. For all models, we do a mean pooling over the final-layer token embedding to aggregate the embeddings to sequence-level. We use an input length of 1,022 amino acids---matching the maximum input length of ESM-1v and the conventional input length for ESM-2 models. For sequences exceeding this length, we split the sequence into overlapping chunks of length 1,022 and aggregate their corresponding embeddings by mean pooling.

\paragraph{Downstream Task} To predict protein fitness, we fit a logistic regression with $L2$ regularization on the different data types (tabular features, embeddings, their concatenation and embeddings after the orthogonal projection). Although a more expressive downstream classifier might increase predictive performance, we deliberately restrict model complexity because we are interested in what is already encoded in the embeddings and not what a classification head might learn. 

\subsection{Metrics}
For predictive performance, we use Matthews correlation coefficient (MCC) to account for imbalances in the data. It ranges from $-$1 to 1 with 1 corresponding to a perfect classifier, 0 to a random or constant classifier, and $-$1 to an inverted perfect classifier. The area under the receiver operating characteristic curve (AUC) is used to choose the degree of regularization. 

\section{Results}

\subsection{Performance Comparison} \label{sec_results_performance}

\begin{figure}
  \centering
  \includegraphics[width=\linewidth]{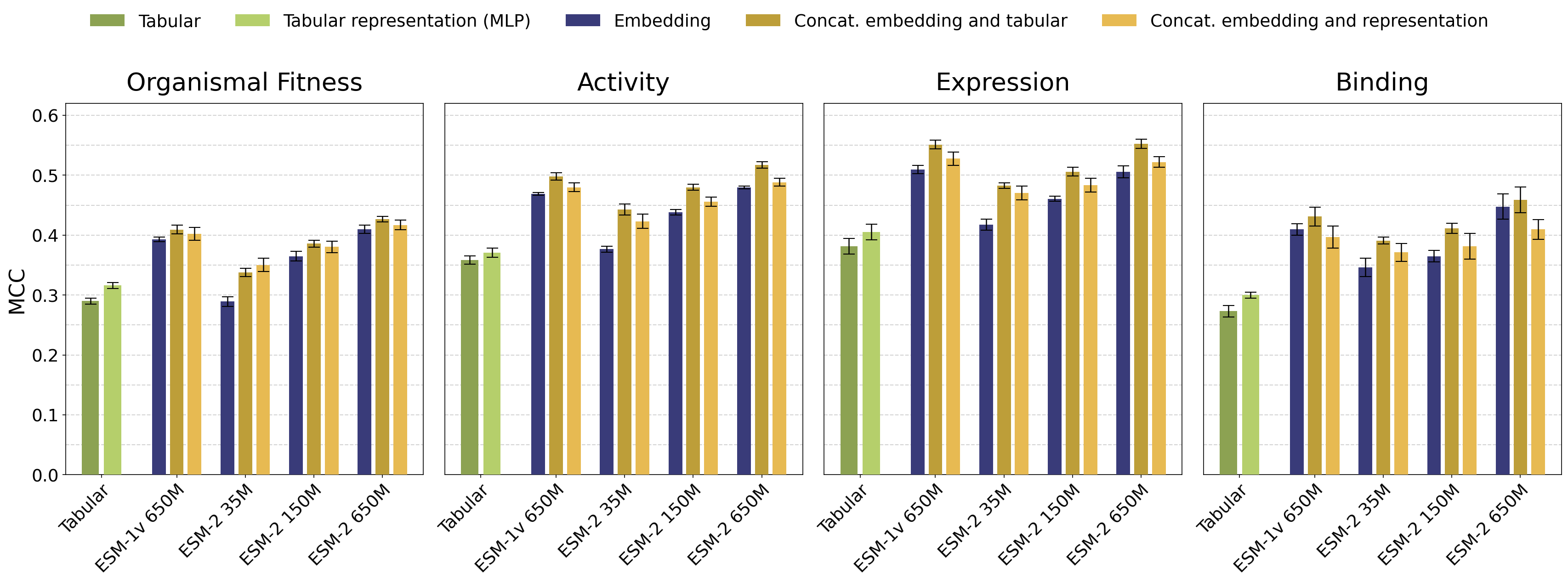}
    \caption{
      Predictive performance measured in Matthews correlation coefficient (MCC) of linear classifiers trained on different data modalities: tabular features (dark green), hidden-layer representation of tabular features (light green), PLM embeddings (blue), and their respective concatenation (dark yellow: embeddings and tabular features, light yellow: embeddings and representation of tabular features) for the four protein functions and PLMs ESM-1v 650M, ESM-2 35M, ESM-2 150M, and ESM-2 650M. The values are averages across folds with corresponding estimated standard deviation as black error bars. 
    }
    \label{fig_mcc_performance}
\end{figure}

Before we analyze the impact of the orthogonal projection, we consider the predictive performance of the linear classifiers fitted on the different data types. Figure \ref{fig_mcc_performance} reports the MCC for models trained on selected tabular features (green), hidden-layer representations of the MLP of selected tabular features (light green), PLM embeddings (blue), and the concatenation of both (dark or light yellow, respectively). For all four protein functions, the embeddings of ESM-1v 650M and ESM-2 650M achieve the highest performance---substantially higher than the tabular features or their representation. Adding the tabular features or their representation to the embeddings leads to only a small or no performance increase, while the largest increase is for the smallest PLM ESM-2 35M. This indicates that the embeddings encode already most of the information provided by the tabular features. For the different ESM-2 models (35M, 150M, and 650M), we observe a slight upward trend in performance when increasing model size. 

In general, the downstream classifiers achieve a higher performance for predicting \textit{Activity} or \textit{Expression} (e.g. MCC of 0.47 and 0.51 for ESM-1v 650M) compared to \textit{Organismal Fitness} or \textit{Binding} (MCC of 0.39 and 0.41 for ESM-1v 650M). The observed performance ranking is stable across the different PLMs and in line with findings of \cite{marquet2024expert,notin2024proteingym}. 
When grouping the binary substitution features by chemical property, we observe very similar results (see Figure \ref{fig_grouped_mcc_performance} in the appendix). This indicates that for the assays and protein functions considered here, it might be sufficient to have less granular information about the amino acid substitution.

For the linear classifier with tabular features only, the number of relevant  features  to predict protein fitness could be considerably reduced via $L1$ regularization while keeping performance within one standard error. Table \ref{table_lasso_results} summarizes the selected features and their percentage of the total number of features for the four protein functions considered. Since many binary features are mutually exclusive, it is not unexpected that on average around 30-50\% of all features obtain a zero coefficient with $L1$ regularization.

For the feature representation of the tabular features, we did not observe a performance increase when scaling up the width and depth of the MLP (not shown here). Therefore, we kept the baseline architecture (two hidden layers and $\hat{p}$ = 256) for the ablation study.

\begin{table}
\caption{Feature selection results using $L1$ regularization for four protein functions. The columns contain the results for  the substitution either on single amino acid level or grouped by chemical property. The number $\widetilde{p}$ denotes average number of features with non-zero coefficients over the five folds with standard deviation in brackets. Additionally, the percentage of the respective total number of features is provided.}
\label{table_lasso_results}
\begin{center}
\begin{tabular}{llrlr}
    \toprule
\multirow{3}{*}{\shortstack[l]{Protein\\Function}}  & \multicolumn{2}{c}{\multirow{2}{*}{\shortstack[l]{Single amino \\ acid features}}}  & \multicolumn{2}{c}{\multirow{2}{*}{\shortstack[l]{Chemical pro-\\perty features}}} \\
& & & & \\
\cmidrule{2-5}
& \multicolumn{1}{c}{$\widetilde{p}$} & \multicolumn{1}{c}{\%} & \multicolumn{1}{c}{$\widetilde{p}$} & \multicolumn{1}{c}{\%} \\
    \midrule
Org. Fitness & 204 (24) & 53 & 53 (1) & 80 \\
Activity & 249 (20) & 64 & 43 (2) & 64 \\
Expression & 260 (10) & 67 & 45 (1) & 67 \\
Binding & 193 (30) & 50 & 30 (6) & 44 \\
    \bottomrule
\end{tabular}
\end{center}
\end{table}

The results of the CITs are reported in Table \ref{tab:pcm-pvalue-all}. For testing for additional information between the embeddings and protein fitness given the tabular features ($H_0:U\perp y\mid X$), we observe $p$-values below the nominal level for 15 out of 16 cases. Hence, except for the embeddings of ESM-2 35M and \textit{Activity}, we would always reject $H_0$ at a Bonferroni-corrected significance level of $\alpha=0.05/8=0.0065$ ($4$ embeddings $\times$ $2$ hypotheses). This means that the embeddings contain statistically significant effects beyond the tabular features across almost all protein functions and models. A plausible reason for the non-significance with the  ESM-2 35M embeddings is the relatively small dimension compared to the other embeddings.
When testing for additional information between the tabular features and protein fitness given the embeddings ($H_0:X\perp y\mid U$), we only observe significant results in four cases: for \textit{Activity} for three models and for \textit{Expression} for the smallest model ESM-2 35M. This indicates that, except for \textit{Activity}, the tabular features contain little to no additional information for protein function beyond the embeddings. This is in line with the observations before when comparing the performance of the embeddings and the concatenations with the tabular features.

\begin{table}
\centering
\sisetup{
  detect-weight = true,
  detect-inline-weight = math,
  text-series-to-math = true,
  group-digits = false,
  exponent-mode = scientific,
  exponent-product = \times,
  table-align-exponent = true,
  mode = text
}
\caption{$p$-values for CITs  for tabular features ($X$) and embeddings ($U$) with protein function ($y$), using $L2$-penalized prediction models within the CIT. Rows are protein language models; columns are protein functions. Each entry is the combination of the $p$-values across folds into a single $p$-value, obtained by $\min(1,\,2\cdot\mathrm{mean}(p))$ \cite{vovk2020combining}. Bold values are significant after Bonferroni correction ($4$ embeddings $\times$ $2$ hypotheses; $\alpha=0.05/8=0.00625$).}
\label{tab:pcm-pvalue-all}
\begin{tabular}{ll S[table-format=1.2e-2] S[table-format=1.2e-2]}
\toprule
PLM & Protein Function & {$U \perp y \mid X$} & {$X \perp y \mid U$} \\
\midrule
\multirow{4}{*}{ESM-1v 650M} & Organismal Fitness & \bfseries 2.91e-7 & 2.54e-1 \\
 & Activity & \bfseries 9.21e-6 & \bfseries 1.59e-3 \\
 & Expression & \bfseries 2.04e-7 & 1.70e-1 \\
 & Binding & \bfseries 5.04e-3 & 7.03e-1 \\
\midrule
\multirow{4}{*}{ESM-2 35M} & Organismal Fitness & \bfseries 8.19e-7 & 7.68e-3 \\
 & Activity & 1.40e-2 & \bfseries 1.20e-3 \\
 & Expression & \bfseries 1.38e-4 & \bfseries 2.77e-11 \\
 & Binding & \bfseries 1.37e-3 & 5.49e-1 \\
\midrule
\multirow{4}{*}{ESM-2 150M} & Organismal Fitness & \bfseries 3.31e-4 & 6.24e-2 \\
 & Activity & \bfseries 6.39e-7 & 1.14e-2 \\
 & Expression & \bfseries 4.74e-6 & 4.87e-2 \\
 & Binding & \bfseries 5.96e-12 & 1.47e-1 \\
\midrule
\multirow{4}{*}{ESM-2 650M} & Organismal Fitness & \bfseries 1.05e-7 & 5.95e-1 \\
 & Activity & \bfseries 6.59e-6 & \bfseries 6.17e-3 \\
 & Expression & \bfseries 8.21e-12 & 4.08e-1 \\
 & Binding & \bfseries 9.38e-7 & 1.00e0 \\
\bottomrule
\end{tabular}
\end{table}

\subsection{Ablation Study}\label{sec_results_ablation}

\begin{figure}
  \centering
  \includegraphics[width=\linewidth]{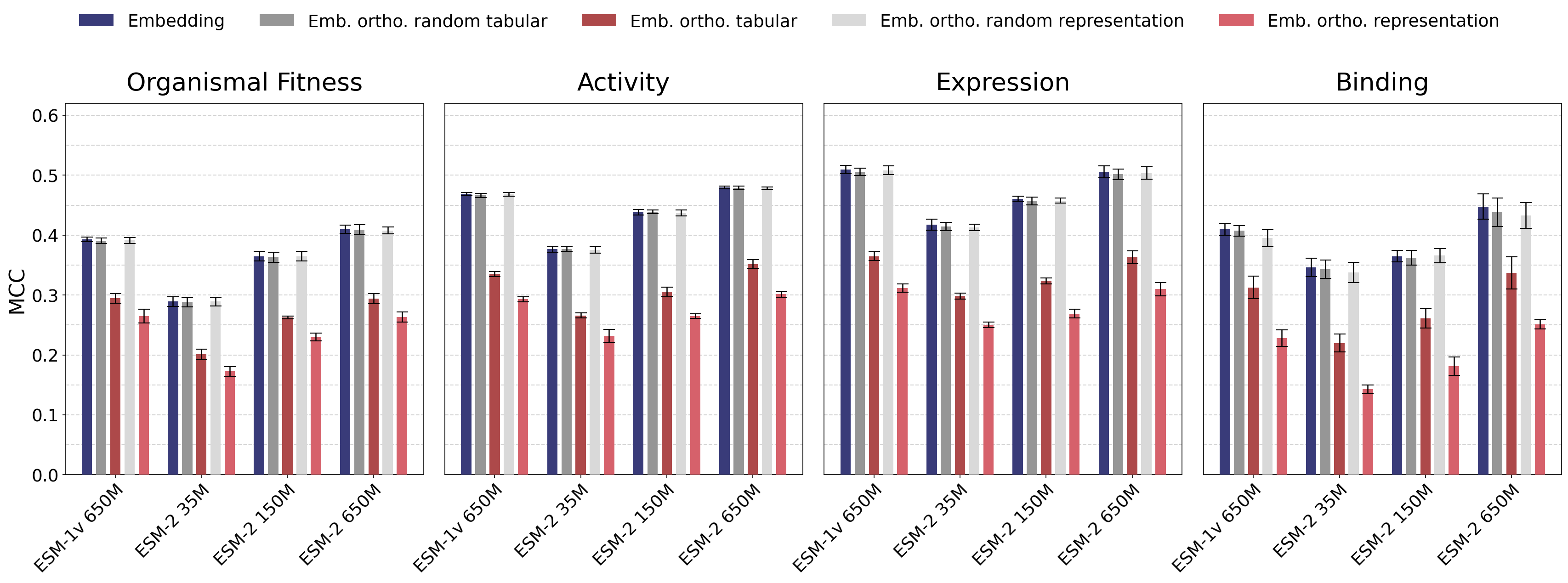}
    \caption{
      Predictive performance measured in Matthews correlation coefficient (MCC) of linear classifiers trained on PLM embeddings (blue) and PLM embeddings orthogonally projected w.r.t. randomized tabular features (dark gray), tabular features (dark red), their randomized representation (light gray) or their representation (light red), respectively, for the four protein functions and PLMs ESM-1v 650M, ESM-2 35M, ESM-2 150M, and ESM-2 650M. The values are averages across folds with corresponding estimated standard deviation as black error bars. For all protein functions and models, the orthogonal projection w.r.t. tabular features (dark red) leads to substantial decrease in predictive performance exceeding the standard deviation observed across folds. If the projection is w.r.t. representation of tabular features (light red), the effect is even stronger. In contrast, for the orthogonal projection w.r.t. to randomized features or representations (dark and light gray), there is only a negligible change in performance.
    }
    \label{fig_across_models_mcc_performance}
\end{figure}

After applying the orthogonal projection to the PLM embeddings with respect to the tabular features or their representation, respectively (red bars in Figure \ref{fig_across_models_mcc_performance}), we see a drop in average performance for all four protein functions and PLMs. 
In particular, the difference in average MCC exceeds the standard deviation observed across folds. For ESM-1v 650M, we see absolute changes in average MCC of $-$0.09 (0.39 $\rightarrow$ 0.30), $-$0.13 (0.47$ \rightarrow$ 0.34), $-$0.14 (0.51$ \rightarrow$ 0.37), and $-$0.10 (0.41 $\rightarrow$ 0.31) for the four protein functions when removing the linear effects of the tabular features.
We observe very similar absolute drops for the three ESM-2 models. 

In all cases, the orthogonal projection with respect to the representation of the tabular features reduces that average performance even further than removing only the linear effects (see light red bar compared to dark red bar in Figure \ref{fig_across_models_mcc_performance}). For ESM-1v 650M, this corresponds to differences in average MCC of $-$0.13, $-$0.18, $-$0.20, and $-$0.18 for the four protein functions.
For the smallest PLM ESM-2 35M, the average MCC falls from 0.35 to 0.15 and hence, the model comes closer to a random classifier. This indicates that higher-order and interaction effects relevant for protein fitness are often already encoded in the embeddings.

For the control experiment with the orthogonal projection for a random subspace, we observe that the average MCC remains essentially unchanged compared to the case without projection (dark and light gray vs. blue in Figure \ref{fig_across_models_mcc_performance}). This supports the claim that the performance drop we observe after the orthogonal projection is specifically due to the removal of patterns within the embeddings correlated with the biochemical features and not due to random removal of information or decreasing the dimension of the embedding space.
The results for the substitution features grouped chemical property are again very similar (see Figure \ref{fig_grouped_across_models_mcc_performance} in the appendix).

Hence, our ablation study shows that the PLM embeddings contain patterns correlated to the effects of the biochemical features which are helpful for protein fitness prediction. Furthermore, our extension of the orthogonal projection to higher-order and interaction effects is capable of removing relevant information beyond linear effects from the embeddings.

\subsection{Evaluation Model}

In our final evaluation, we consider how much of the variance in the embedding-based predictions can be explained with the biochemical features. Further, we analyze how these features correlate with embedding-based predictions and whether this relates to know biochemical properties. For this evaluation model, we regress the embedding-based predicted logits on the tabular features or their representation, respectively, on the test set. The corresponding adjusted $R^2$ is reported in Table \ref{table_eval_r2_results}. We consider here the embeddings of ESM-1v 650M. We observe that both the tabular features and their representation can explain a substantial part of the variance in the embedding-based predicted logits, i.e., ranging from 0.15 to 0.29. Hence, up to 29\% of the variance in the predicted logits can be attributed to the biochemical features. This indicates that the embeddings encode patterns correlated to the tabular features that are relevant for protein fitness prediction. For \textit{Activity} and \textit{Expression}, we observe the highest $R^2$. This is in line with results from section \ref{sec_results_performance}, where we saw the highest performance among classifiers trained on tabular features. For \textit{Binding}, we see the lowest adjusted $R^2$. Again, this is in line with results from the previous section, where the tabular features had the lowest predictive performance. After applying the randomized orthogonal projection to the embeddings, the $R^2$ remains essentially unchanged as we have seen with the performance in the previous section. For the regular (non-randomized) orthogonal projection, the $R^2$ drops close to zero. This is expected because after removing any linear effects of the tabular features or their representation, respectively, they cannot explain any variance in the embedding-based predicted logits anymore. As this evaluation is performed on the test set, the value is not expected to be exactly zero. In particular, when fitting the evaluation model on the same data on which the orthogonal projection was calculated, we see a drop to exactly zero (omitted here). For the different ESM-2 models, we observe similar adjusted $R^2$ (see Table \ref{table_eval_r2_results_esm2} in the appendix) compared to the ESM-1v embeddings. 

\begin{table}
\centering
\fontsize{8pt}{9.5pt}\selectfont
\caption{Results of evaluation model for four protein functions: Adjusted $R^2$ for regressing embedding-based predicted logits (ESM-1v 650M) on tabular features or representation of tabular features, respectively, for embeddings without orthogonal projection (`No ortho'), after applying the randomized orthogonal projection (`Rand. ortho.'), and after (regular) orthogonal projection (`Ortho.'). The values are averages across folds with corresponding estimated standard deviation in brackets.}
\label{table_eval_r2_results}
\begin{tabular}{llrlrlr}
    \toprule
\multirow{3}{*}{Protein Function}  & \multicolumn{6}{c}{Adjusted $R^2$} \\
    \cmidrule{2-7}
& \multicolumn{3}{c}{Tabular} & \multicolumn{3}{c}{Repr. Tab.} \\
    \cmidrule{2-7}
& \multicolumn{1}{c}{No ortho.} & \multicolumn{1}{c}{Rand. ortho.} & \multicolumn{1}{c}{Ortho.} & \multicolumn{1}{c}{No ortho.} & \multicolumn{1}{c}{Rand. ortho.} & \multicolumn{1}{c}{Ortho.}\\
    \midrule
Org. Fitness & 0.189 (0.006) & 0.189 (0.006) & 0.002 (0.001) & 0.234 (0.005) & 0.234 (0.005) & 0.007 (0.001) \\
Activity & 0.220 (0.006) & 0.221 (0.005) & 0.004 (0.002) & 0.250 (0.005) & 0.251 (0.004) & 0.006 (0.003) \\
Expression & 0.262 (0.004) & 0.262 (0.004) & 0.008 (0.003) & 0.289 (0.012) & 0.288 (0.012) & 0.009 (0.003) \\
Binding & 0.153 (0.024) & 0.153 (0.021) & 0.018 (0.009) & 0.206 (0.014) & 0.203 (0.008) & 0.011 (0.005) \\
    \bottomrule
\end{tabular}
\end{table}

\sisetup{
  mode = text
}
\begin{table*}[t]
\caption{Coefficients of structural features of evaluation model for four protein functions: average coefficients across folds for regressing embedding-based predicted logits (ESM-1v 650M) on structural tabular features. Significance refers to whether estimate was significant across all folds. Blanks refer to variables that were excluded during features selection.}
\label{table_g2p_coeffs}
\begin{center}

\begin{tabular}{lSSSS}
    \toprule
\multirow{2}{*}{Feature}  & \multicolumn{4}{c}{Protein Function} \\
    \cmidrule{2-5}
& \multicolumn{1}{c}{Org. Fitness} & \multicolumn{1}{c}{Activity} & \multicolumn{1}{c}{Expression} & \multicolumn{1}{c}{Binding} \\
    \midrule
\textit{phi} & -0.051\,* & -0.021\, & -0.127\,* & -0.145\,* \\
\textit{psi} & 0.003\, & 0.052\, & -0.011\, & 0.010\, \\
\textit{plddt} & -0.139\,* & -0.138\,* & -0.117\,* & -0.097\, \\
\textit{rsa} & 0.316\,* & 0.425\,* & 0.467\,* & 0.339\,* \\
\textit{strand} & \multicolumn{1}{c}{---} & 0.026\, & -0.274\, & -0.606\,* \\
\textit{coil} & 0.005\, & 0.016\, & 0.094\, & \multicolumn{1}{c}{---} \\
\textit{helix} & 0.280\,* & 0.010\, & \multicolumn{1}{c}{---} & 0.232\, \\
    \bottomrule
    \multicolumn{5}{c}{*: Significant after the Benjamini–Hochberg procedure at FDR=5\%}
\end{tabular}

\end{center}

\end{table*}

As a last step of our evaluation, we consider the estimated coefficients of the evaluation model. In particular, we are interested in which features are correlated with the embedding-based predictions and whether these correlations correspond to known biological properties. As for the $R^2$, we present here the results for the embeddings of ESM-1v. However, we observe very similar coefficients for the ESM-2 650M embeddings (see Table \ref{table_g2p_coeffs_esm2} and Figure \ref{fig_coeff_heatmap_esm2} in the appendix). The coefficients for the structural features and whether they were significant in all folds are reported in Table \ref{table_g2p_coeffs}. 
For the values of coefficients, we see shared patterns across protein function. For example, relative solvent accessibility (RSA), as a measure of how exposed or buried an amino acid residue is within a protein's 3D structure, contributes positively to prediction for all protein functions. It is even statistically significant across folds for all protein functions. This is in line with greater surface exposure of the affected amino acid being associated with reduced protein disruption. Consistently, the plddt score reflecting confidence in structure prediction is negatively correlated with the embeddings predicting protein fitness. This is in line with the score being related to protein flexibility and thus, functionally relevant regions \cite{vander2025flexibility}.

\begin{figure}
  \centering
  \includegraphics[width=\linewidth]{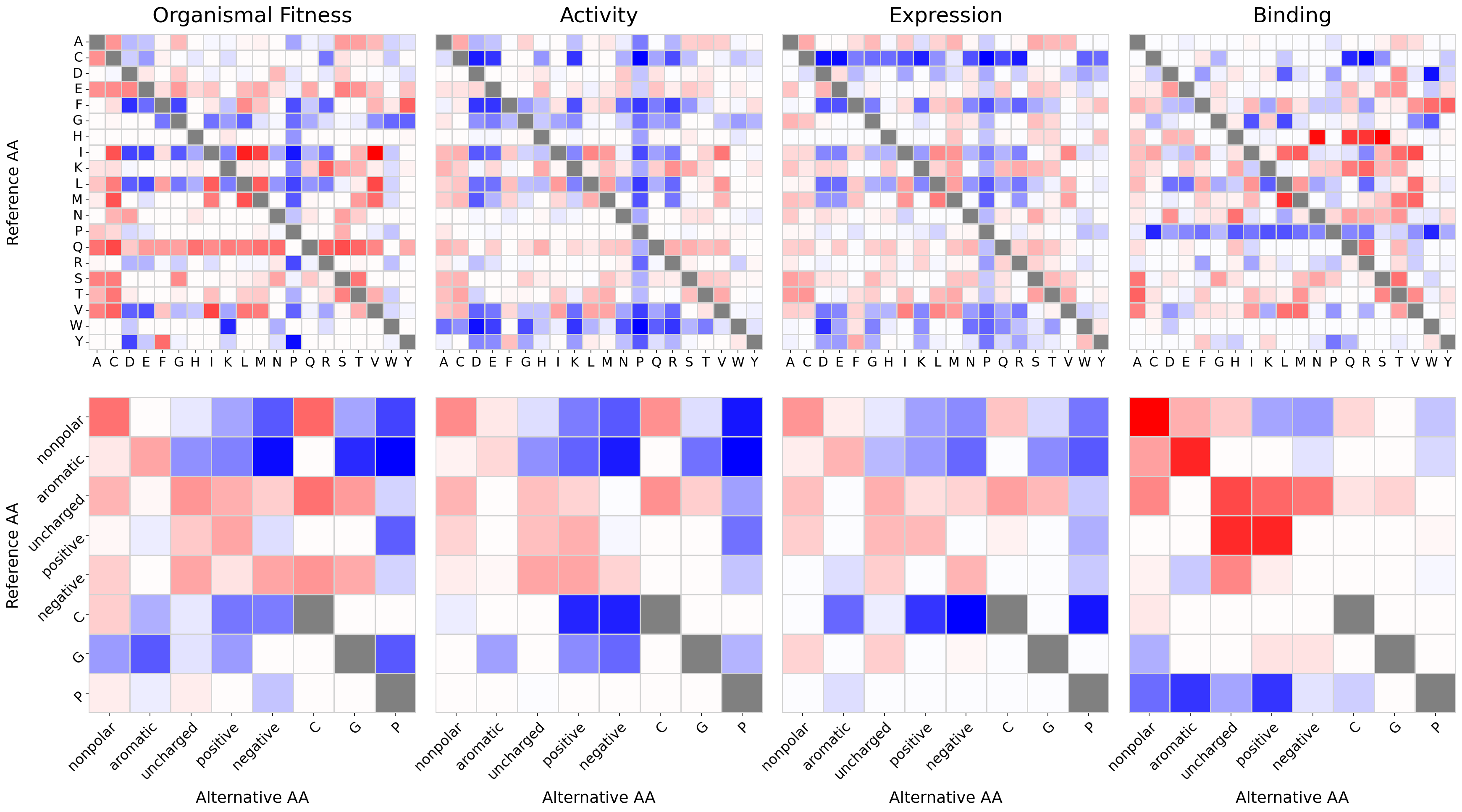}
    \caption{
      Heatmaps of coefficients of (binary) substitution features of evaluation model, i.e., fitted on the embedding-based predictions (ESM-1v 650M), for different protein functions. The reference amino acid (row) of the wild-type protein was substituted for the alternative amino acid (column). Upper plots illustrate coefficients for single amino acid substitution features, lower ones for substitution features grouped by chemical property. Every entry corresponds to coefficient with either positive (red) or negative (blue) contribution to the predicting the positive class. Coefficients of features excluded by the feature selection during training are set to zero (white). The values of coefficients are averages across folds.
    }
    \label{fig_coeff_heatmap_esm1}
\end{figure}

The coefficients for the binary substitution features are illustrated as heatmaps in Figure \ref{fig_coeff_heatmap_esm1}. They display several patterns. 
For example, for \textit{Organismal Fitness}, \textit{Activity}, and \textit{Expression}, a substitution of almost every reference amino acid for Proline (P) correlates with a lower predicted logit of the embedding-based classifier. Proline is known to have distinctive role for protein structure because of its rigidity, i.e., it usually ends a helix or introduces a kink \cite{cordes2002proline}. Hence, a substitution can be disruptive for protein fitness. In contrast, substituting Glutamine (Q) contributes mostly positively to the predicted logits for \textit{Organismal Fitness}. It is a more flexible amino acid and not necessarily essential for maintaining secondary structure. When considering the substitution features grouped by chemical property (lower row in Figure \ref{fig_coeff_heatmap_esm1}, we can also recover expected protein behavior. For example, for \textit{Organismal Fitness}, \textit{Activity}, and \textit{Expression}, a substitution of an aromatic amino acid for a non-aromatic correlates negativity with the embedding-based classifier predicting protein fitness. Aromatic amino acids are known to contribute to protein structure due to interactions involving their aromatic rings \cite{calinsky2024aromatic}. 

\section{Conclusion}

In this work, we presented a computationally efficient approach to interpret embeddings of PLMs. In particular, we extended and applied an orthogonal projection technique to remove biochemical features from the embeddings. We could show that this led to a substantial decrease in performance of a classifier trained on these embeddings for protein fitness prediction. Furthermore, our additional control experiments---where the orthogonal projection was calculated with respect to a random subspace---showed that the decrease is not due to random removal of information or dimensionality reduction but actually driven by the removal of relevant biochemical features. In a further evaluation, we saw that these biochemical and their higher-order and interaction effects actually explain part of the variance in the predictions of the embedding-based classifier. These findings confirm previous results on PLM embeddings and predictive biochemical features for protein fitness prediction. This approach is not limited to the features considered in this analysis and easily transferable to a broader class of biological features. In future work, we also aim to investigate how these results generalize to other PLMs beyond the ESM family and PLMs fine-tuned for specific protein functions. As PLMs can help predict the effect of a variant on protein function, they can also contribute to understanding the underlying biochemical mechanism of many diseases. Furthermore, they have the potential to advance drug design, which in biomedical science, often involves targeting a specific protein or its function. For all of these purposes, approaches that provide interpretability are needed.

\subsubsection*{Acknowledgements}
We gratefully acknowledge funding by grants KL 3037/7-1 (to NK) and RE 3474/8-1 (to BYR), project P5 in the Research Unit KI-FOR 5363 (grant 459422098) of the German Research Foundation (DFG).

\subsubsection*{Disclosure of Interests}
The authors have no competing interests to declare that are
relevant to the content of this article.

\subsubsection*{Use of Generative AI}
The authors used ChatGPT by OpenAI and Claude by Anthropic for language editing of portions of the manuscript and for assistance with Python code development. All AI-generated content was carefully reviewed and edited by the authors, who take full responsibility for the accuracy, originality, and integrity of the manuscript. Generative AI was not used for data analysis, interpretation of results, or generation of scientific conclusions.

\bibliographystyle{splncs04}
\bibliography{references}

\newpage
\appendix

\section*{Appendix}

\begin{figure}[h]
  \centering
  \includegraphics[width=\linewidth]{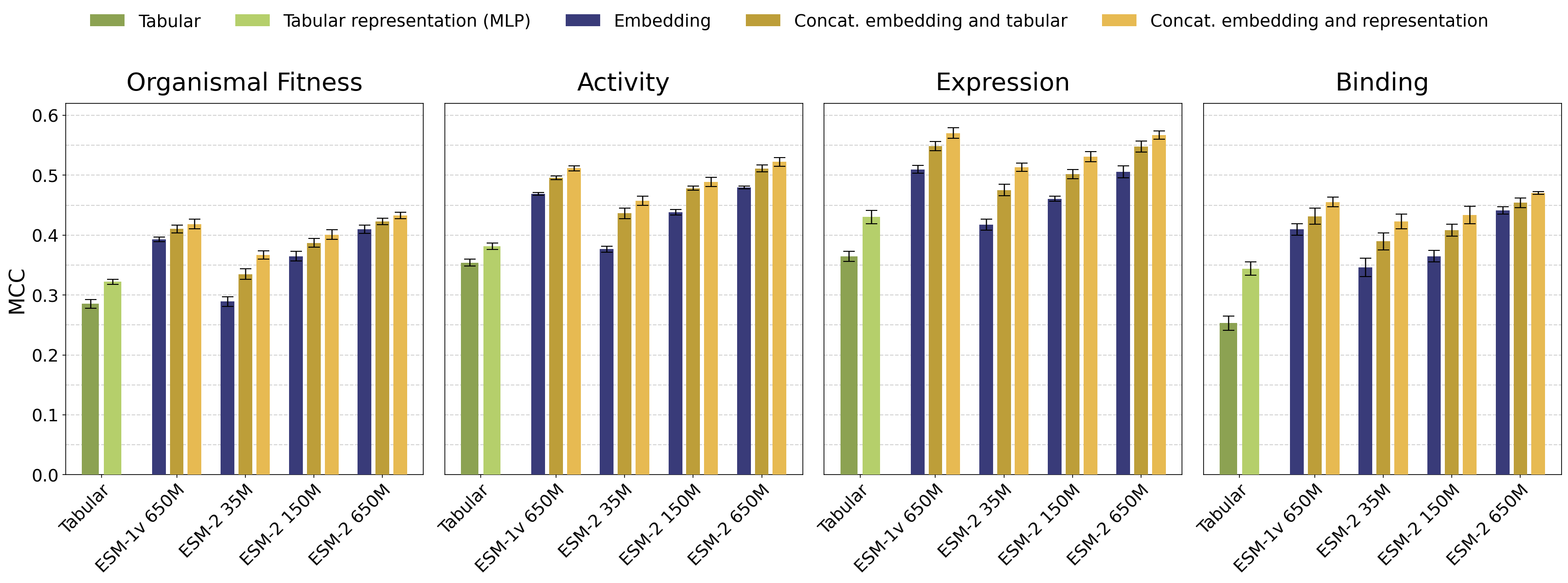}
    \caption{
      Predictive performance measured in Matthews correlation coefficient (MCC) of linear classifiers trained on different data modalities: tabular features where the substitution features where grouped by biochemical property (dark green), hidden-layer representation of tabular features (light green), PLM embeddings (blue), and their respective concatenation (dark yellow: embeddings and tabular features, light yellow: embeddings and representation of tabular features) for the four protein functions and PLMs ESM-1v 650M, ESM-2 35M, ESM-2 150M, and ESM-2 650M. The values are averages across folds with corresponding estimated standard deviation as black error bars.
    }
    \label{fig_grouped_mcc_performance}
\end{figure}

\begin{figure}[h]
  \centering
  \includegraphics[width=\linewidth]{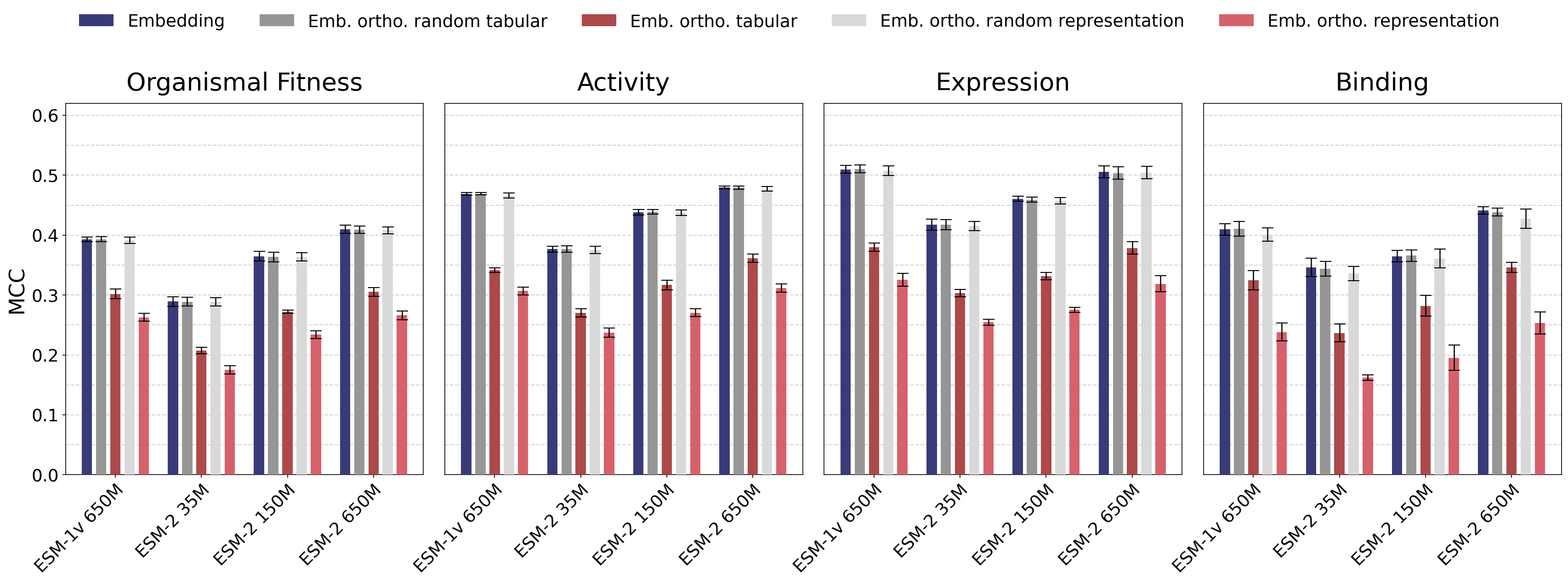}
    \caption{
    Predictive performance measured in Matthews correlation coefficient (MCC) of linear classifiers trained on PLM embeddings (blue) and PLM embeddings orthogonally projected w.r.t. randomized tabular features (dark gray), tabular features (dark red), their randomized representation (light gray) or their representation (light red), respectively, for the four protein functions and PLMs ESM-1v 650M, ESM-2 35M, ESM-2 150M, and ESM-2 650M. The substitution features of the tabular features are grouped by biochemical property. The values are averages across folds with corresponding estimated standard deviation as black error bars. For all protein functions and models, the orthogonal projection w.r.t. tabular features (dark red) leads to substantial decrease in predictive performance exceeding the standard deviation observed across folds. If the projection is w.r.t. representation of tabular features (light red), the effect is even stronger. In contrast, for the orthogonal projection w.r.t. to randomized features or representations (dark and light gray), there is only a negligible change in performance.
    }
    \label{fig_grouped_across_models_mcc_performance}
\end{figure}

\clearpage

\begin{table}
\centering
\fontsize{8pt}{9.5pt}\selectfont
\caption{Results of evaluation model for four protein functions (`PF'): Adjusted $R^2$ for regressing embedding-based predicted logits (ESM-2 35M, 150M, and 650M) on tabular features or representation of tabular features, respectively, for embeddings without orthogonal projection (`No ortho'), after applying the randomized orthogonal projection (`Rand. ortho.'), and after (regular) orthogonal projection (`Ortho.'). The values are averages across folds with corresponding estimated standard deviation in brackets.}
\label{table_eval_r2_results_esm2}
\begin{tabular}{lllrlrlr}
    \toprule
\multirow{3}{*}{\shortstack[l]{ESM-2\\size}}  & \multirow{3}{*}{PF}  & \multicolumn{6}{c}{Adjusted $R^2$} \\
    \cmidrule{3-8}
& & \multicolumn{3}{c}{Tabular} & \multicolumn{3}{c}{Repr. Tab.} \\
    \cmidrule{3-8}
& & \multicolumn{1}{c}{No ortho.} & \multicolumn{1}{c}{Rand. ortho.} & \multicolumn{1}{c}{Ortho.} & \multicolumn{1}{c}{No ortho.} & \multicolumn{1}{c}{Rand. ortho.} & \multicolumn{1}{c}{Ortho.} \\
    \midrule
\multirow{4}{*}{35M} & Org. & 0.170 (0.007) & 0.169 (0.007) & 0.001 (0.001) & 0.189 (0.004) & 0.190 (0.004) & 0.004 (0.001) \\
& Act. & 0.217 (0.008) & 0.217 (0.008) & 0.006 (0.002) & 0.224 (0.008) & 0.224 (0.008) & 0.006 (0.002) \\
& Expr. & 0.212 (0.006) & 0.211 (0.006) & 0.010 (0.004) & 0.220 (0.007) & 0.218 (0.007) & 0.008 (0.003) \\
& Bind. & 0.157 (0.017) & 0.156 (0.017) & 0.010 (0.011) & 0.210 (0.020) & 0.208 (0.021) & 0.014 (0.002) \\
    \midrule
\multirow{4}{*}{150M} & Org. & 0.188 (0.005) & 0.189 (0.003) & 0.003 (0.001) & 0.238 (0.004) & 0.239 (0.006) & 0.006 (0.001) \\
& Act. & 0.223 (0.008) & 0.225 (0.008) & 0.004 (0.002) & 0.251 (0.006) & 0.253 (0.006) & 0.008 (0.002) \\
& Expr. & 0.241 (0.007) & 0.241 (0.007) & 0.007 (0.004) & 0.266 (0.007) & 0.265 (0.007) & 0.011 (0.003) \\
& Bind. & 0.150 (0.015) & 0.147 (0.015) & 0.009 (0.006) & 0.200 (0.010) & 0.196 (0.010) & 0.008 (0.012) \\
    \midrule
\multirow{4}{*}{650M} & Org. & 0.176 (0.004) & 0.176 (0.004) & 0.001 (0.001) & 0.231 (0.003) & 0.231 (0.003) & 0.007 (0.001) \\
& Act. & 0.209 (0.010) & 0.210 (0.010) & 0.006 (0.002) & 0.242 (0.007) & 0.243 (0.006) & 0.006 (0.002) \\
& Expr. & 0.230 (0.008) & 0.229 (0.007) & 0.008 (0.003) & 0.242 (0.010) & 0.241 (0.010) & 0.011 (0.001) \\
& Bind. & 0.167 (0.018) & 0.164 (0.018) & 0.015 (0.010) & 0.221 (0.016) & 0.219 (0.010) & 0.008 (0.006) \\
    \bottomrule
\end{tabular}
\end{table}

\begin{table*}[t]
\caption{Coefficients of structural features of evaluation model for four protein functions: average coefficients across folds for regressing embedding-based predicted logits (ESM-2 650M) on structural tabular features. Significance refers to whether estimate was significant across all folds. Blanks refer to variables that were excluded during features selection.}
\label{table_g2p_coeffs_esm2}
\begin{center}

\begin{tabular}{lSSSS}
    \toprule
\multirow{2}{*}{Feature}  & \multicolumn{4}{c}{Protein Function} \\
    \cmidrule{2-5}
& \multicolumn{1}{c}{Org. Fitness} & \multicolumn{1}{c}{Activity} & \multicolumn{1}{c}{Expression} & \multicolumn{1}{c}{Binding} \\
    \midrule
\textit{phi} & -0.052\,** & \multicolumn{1}{c}{---} & -0.125\,* & -0.082\, \\
\textit{psi} & 0.002\, & 0.060\, & \multicolumn{1}{c}{---} & -0.023\, \\
\textit{plddt} & -0.134\,* & -0.138\,* & -0.061\, & -0.132\, \\
\textit{rsa} & 0.348\,* & 0.395\,* & 0.494\,* & 0.415\,* \\
\textit{strand} & \multicolumn{1}{c}{---} & -0.096\, & -0.298\, & -0.427\, \\
\textit{coil} & -0.052\, & 0.031\, & 0.086\, & \multicolumn{1}{c}{---} \\
\textit{helix} & 0.230\,* & 0.065\, & \multicolumn{1}{c}{---} & 0.335\, \\
    \bottomrule
    \multicolumn{5}{c}{*: Significant after the Benjamini–Hochberg procedure at FDR=5\%}
\end{tabular}

\end{center}

\end{table*}

\begin{figure}
  \centering
  \includegraphics[width=\linewidth]{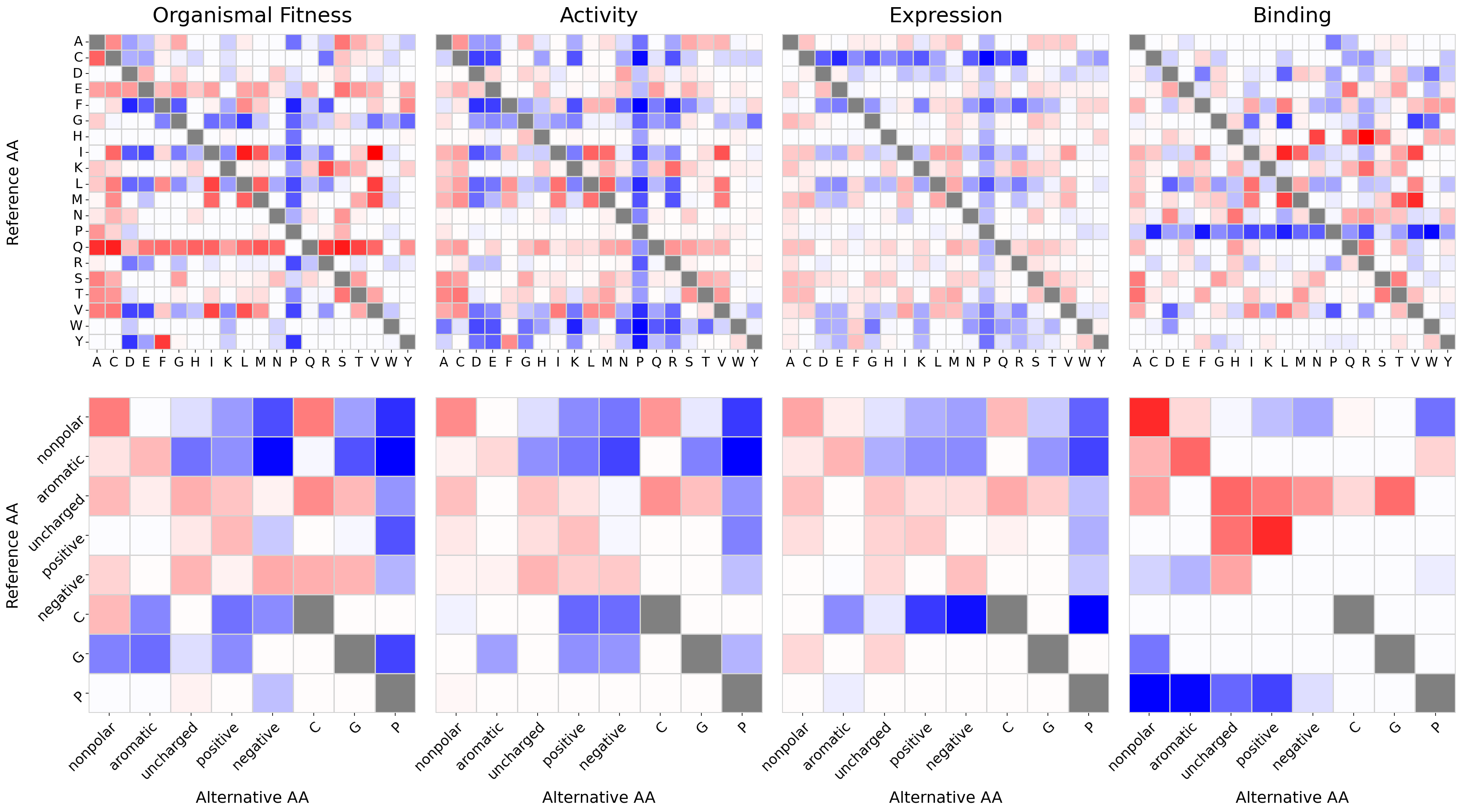}
    \caption{
      Heatmaps of coefficients of (binary) substitution features of evaluation model, i.e., fitted on the embedding-based predictions (ESM-2 650M), for different protein functions. The reference amino acid (row) of the wild-type protein was substituted for the alternative amino acid (column). Upper plots illustrate coefficients for single amino acid substitution features, lower ones for substitution features grouped by chemical property. Every entry corresponds to coefficient with either positive (red) or negative (blue) contribution to the predicting the positive class. Coefficients of features excluded by the feature selection during training are set to zero (white). The values of coefficients are averages across folds.
    }
    \label{fig_coeff_heatmap_esm2}
\end{figure}

\end{document}